\relax
\documentclass[letterpaper]{article} 
\usepackage{aaai22}  
\usepackage{times}  
\usepackage{helvet}  
\usepackage{courier}  
\usepackage[hyphens]{url}  
\usepackage{graphicx} 
\usepackage{amsfonts}
\usepackage{todonotes}
\usepackage{natbib}  
\usepackage{caption} 
\DeclareCaptionStyle{ruled}{labelfont=normalfont,labelsep=colon,strut=off} 
\title{Text Style Transfer for Bias Mitigation \\ using Masked Language Modeling}
\author {
    Ewoenam Kwaku Tokpo, 
    Toon Calders 
}
\affiliations {
    \ Department of Computer Science, University of Antwerp\\
}

\begin{document}

\maketitle

\begin{abstract}
It is well known that textual data on the internet and other digital platforms contain significant levels of bias and stereotypes. Although many such texts contain stereotypes and biases that inherently exist in natural language for reasons that are not necessarily malicious, there are crucial reasons to mitigate these biases. For one, these texts are being used as training corpus to train language models for salient applications like cv-screening, search engines, and chatbots; such applications are turning out to produce discriminatory results. Also, several research findings have concluded that biased texts have significant effects on the target demographic groups. For instance, masculine-worded job advertisements tend to be less appealing to female applicants.
  
In this paper, we present a text style transfer model that can be used to automatically debias textual data. Our style transfer model improves on the limitations of many existing style transfer techniques such as loss of content information. Our model solves such issues by combining latent content encoding with explicit keyword replacement. We will show that this technique produces better content preservation whilst maintaining good style transfer accuracy.
\end{abstract}

\section{Introduction}
\subsection{Bias in textual data}

Over years, Natural Language Processing (NLP) models have evolved from elementary techniques like rule-based systems \cite{brill1992simple} to more complex and powerful models and architectures like the Transformer \cite{vaswani2017attention} which has spun off models such as BERT \cite{devlin2018bert}, GPT-3 \cite{NEURIPS2020_1457c0d6} and XLNET \cite{yang2019xlnet}. The application domain of NLP has also evolved greatly over the years; from
search engines and chatbots to CV-screening applications. Accordingly, the issue of fairness has become of great interest in NLP. Authors such as \citet{bolukbasi2016man} and \citet{may2019measuring} have drawn attention to some fairness problems in the domain. In a post on BuzzFeed \cite{BuzzFeed} with the title, ``Scientists Taught A Robot Language. It Immediately Turned Racist", the author reports how various automated language systems are disturbingly learning discriminatory patterns from data. Another prominent example is Amazon's AI recruitment tool which turned out to be biased against female applicants \cite{reuters}.  As such, there is a strong need to find effective ways of training fair models. Debiasing textual data before training can be an important preprocessing step in training fair language systems like chatbots, language translation systems, and search engines.

A more direct need for debiasing textual data has been pointed out
by various researchers \cite{gaucher2011evidence,tang2017gender,hodel2017gender} who have uncovered the worrying issue of bias in job advertisements. This can have significant implications on the job recruitment process. As a matter of fact, \citet{gaucher2011evidence} explored the effect of biased job advertisements on participants of a survey. 
They found that changing the wording of a job advertisement to favor a particular gender group considerably reduced the appeal of the job to applicants not belonging to that gender, regardless of the gender stereotype traditionally associated with the job.
Consequent to such findings, a few tools and models have been developed to detect and mitigate biases in job advertisements. Some of these tools include text editors like Textio which has been successfully used by companies such as Atlassian to increase diversity in their workforce \cite{daugherty2019using}. 

Another area of impact, regarding biased texts, is in news publications; \citet{kiesel2019semeval} explore the issue of hyperpartisan news from an extreme left or right-wing perspective. Again, with the prevalence of hate speech and microaggression perpetuated on various social media platforms, there have been growing concerns about fairness in such areas. 

\subsection{Style transfer}
A machine learning technique that can be employed to debias text documents is style transfer. Style transfer is a technique that involves converting text or image instances from one domain to another, such that the content and meaning of the instance largely remain the same but the style changes. Applying style transfer specifically as a text debiasing technique has not been significantly explored although there has been research done on closely-related applications like moderating hate-speech \cite{santos2018fighting} in text, and gender obfuscation \cite{reddy2016obfuscating}, where the objective is to disguise the gender of an author of a text as a means of privacy protection or for the prevention of inadvertent discrimination of the author. 

A problem that has challenged research in text style transfer is the relative unavailability of parallel data that would ideally be required to train such models \cite{rao2018dear,fu2018style,shen2017style}. Training with parallel data makes it possible to directly map training instances from one domain to the other, hence, facilitating the learning process. Another challenge with text style transfer is the non-differentiability of discrete words which causes optimization problems \cite{yang2018unsupervised}.
Due to the mentioned issues, most style transfer systems mainly employ training techniques that fall under two categories, keyword replacement, and Auto-encoder sequence-to-sequence techniques. As we will discuss later, these two techniques have major challenges such as significant loss of content information and lack of fluency. In the case of keyword replacement, biased words are deleted and replaced with alternative words; leading to complete loss of information from the deleted words \cite{sudhakar2019transforming}, whereas, in the case of the Auto-encoder sequence-to-sequence generative approach, the input text is directly encoded by an encoder to get a latent representation of the text, which is subsequently decoded by a decoder. As a result, significant information is lost and the decoder is also unable to properly reconstruct the text fluently. 

Given the female-biased text, \textit{``The event was kid-friendly for all the \textbf{mothers} working in the company"}, our task is to transform this text into a gender-neutral version like \textit{``The event was kid-friendly for all the \textbf{parents} working in the company"}.
In this paper, we present a style transfer model that performs such tasks. In addition to this, our model improves the above-mentioned limitations of both the Auto-encoder sequence-to-sequence technique and the Explicit Style Keyword Replacement technique whilst maintaining a good style transfer accuracy. 
This we achieve by combining key components of both techniques such that they complement each other;
the fact that Explicit Style Keyword Replacement models retain sufficient content information by not compressing the whole text into a latent representation, whereas Auto-encoder sequence-to-sequence models also retain information by not deleting tokens. Our model adds latent content representation to the Keyword replacement process. To the best of our knowledge, we are the first to use this technique for text style transfer. Furthermore, our work introduces some key novelties. Firstly, we propose a new and more effective approach for identifying attribute tokens (words that contribute to bias in a text) by using a text classification explanation model. Secondly, we present a unique way of generating latent content representation using a dual objective to train a token embedder. The details of these techniques will be discussed in subsequent sections.

To summarize, the main contributions of this work include:
\begin{enumerate}
    \item The development of an end-to-end text debiasing model that can convert a piece of biased text to a neutral version whilst maintaining significant content information. Our model is trained exclusively on nonparallel data; since parallel corpora are relatively hard to obtain, training with only nonparallel data is of great importance.
 
    \item A novel way of improving content preservation and fluency in text style transfer; by combining keyword replacement and latent content information. We will show that this significantly improves content preservation in style transfer whilst ensuring good style transfer accuracy. Some other key novelties in our work include our approach to generating latent content representation and our approach to identifying attribute tokens.  
\end{enumerate}

We make the code and data used in this work available \footnote{https://github.com/EwoeT/MLM-style-transfer}.

\section{Related work}

Style transfer has been widely and successfully explored in computer vision to convert images from one style to the other \cite{gatys2016image,huang2017arbitrary,johnson2016perceptual}; for instance, \citet{gatys2016image} show how the style of a photograph can be converted to the style of an artwork by combining the content of the photograph and the style of the painting. However, directly applying image style transfer techniques for text is problematic because of the unique characteristics of both domains \cite{hu2020text}.
In NLP, style transfer has mostly been explored in areas such as sentiment analysis \cite{li2018delete,fu2018style,zhang2018style} and machine translation \cite{lample2017unsupervised}.

A few style transfer learning techniques use parallel data for training.
\citet{hu2020text} give an elaborate survey on such models. In this paper, we will only focus on models that are trained on non-parallel data, some of which we will review in the following subsection.

\subsection{Auto-encoder sequence-to-sequence models}

Auto-encoder sequence-to-sequence models basically consist of an encoder that encodes the given text into a latent representation which is then decoded by a decoder. 
Many of these models adopt an adversarial approach to learn to remove any style attribute from the latent representation. The resulting disentangled latent representation is decoded by the decoder in a sequential generative manner.

\subsubsection{Aligned auto-encoder model and Cross-aligned auto-encoder model: } \citet{shen2017style} propose two models for text style transfer based on the Auto-encoder sequence-to-sequence technique; an aligned auto-encoder model and a variant of that called the cross-aligned auto-encoder model. The aligned auto-encoder model uses an adversarial approach to teach the encoder to generate a disentangled latent representation that can be decoded into the target text. The cross-aligned variant bases on the idea that the transferred source representation should exhibit the same distribution as the target representation and vice versa. Two discriminators are used in the adversarial training, one to distinguish a real target representation from a transferred source representation and the other to distinguish a real source representation from a transferred target representation. 

\subsubsection{Back-translation style transfer model (BST): } \citet{prabhumoye2018style} propose a style transfer model using back-translation. This is based on prior research that suggests that language translation retains the meaning of a text but not the stylistic features \cite{rabinovich-etal-2017-personalized}. Hence, they encode a disentangled latent representation of the input text using back-translation. 
A bidirectional LSTM is then used to decode the disentangled latent representation into a text in the target style. They use adversarial learning to ensure that the generated text belongs to the target style. Because of the discrete and non-differentiable nature of tokens, they approximate discrete training by using a continuous approximation of softmax values to sample tokens.

\subsubsection{}
An issue with Auto-encoder sequence-to-sequence models, in general, is the loss of information due to compression when encoding. Furthermore, \citet{wu2019mask} note that sequence-to-sequence models for style transfer often have limited abilities to produce high-quality hidden representation and are unable to generate long meaningful sentences.
Nonetheless, sequence-to-sequence generative models can prove more effective in applications where the text needs to be considerably rephrased (eg. informal style to a formal style).

\subsection{Explicit Style Keyword Replacement}

These methods follow the general approach of identifying attribute markers, deleting these markers, and predicting appropriate replacements for these markers which conform to the target style. 

\subsubsection{DeleteOnly model and Delete\&Retrieve model: } 
\citet{li2018delete} propose the DeleteOnly and the Delete\&Retrieve, which use a three-step Delete, Retrieve, and Generate approach. In the Delete phase, attribute tokens are identified and removed from the input text based on their relative frequencies in the respective style domains.
In the Retrieve stage, possible replacement words are predicted based on similar sentences in the target style. For the Generate stage, both methods use an RNN to encode context tokens (other words aside attribute tokens in the input text) and decode them to the target style.
In the DeleteOnly method, the RNN takes as inputs only the context tokens and the target style embedding to generate the target sentence, whereas the Delete\&Retrieve method takes as inputs the context tokens, the target style embedding, and the replacement words from the Retrieve stage to construct a target sentence.

\subsubsection{B-GST and G-GST: } \citet{sudhakar2019transforming} introduce Blind Generative Style Transformer (B-GST) and Guided Generative Style Transformer (G-GST) as improvements on DeleteOnly and the Delete\&Retrieve from  \cite{li2018delete}. They use attention weights from a transformer-based style classifier to identify attribute words. The Retrieve phase generally follows the same approach as \cite{li2018delete}. For the generate phase, they implement their version of DeleteOnly and the Delete\&Retrieve with GPT \cite{radford2018improving}, and refer to them as  B-GST and G-GST respectively.

\subsubsection{}
Since Explicit Style Keyword Replacement methods only delete a small portion of the input text, they preserve much more information. These systems on the other hand are unable to properly capture information of the deleted tokens \cite{sudhakar2019transforming}, 
leading to examples such as \textit{``The event was kid-friendly for all the \textbf{mothers} working in the company"} 
$\rightarrow$ \textit{``The event was kid-friendly for all the \textbf{children} working in the company"}.

\section{Methodology}
The goal of our model is to transform any piece of biased text into a neutral version. 
To put this formally, we take the two style attributes  $s_a$ and $s_b$ to represent neutral style and biased style respectively. Given a text corpus $X_b = \{x_{b_1},...,x_{b_N}\} $ that belongs to $s_b$, 
our goal is to convert a given text $x_b$ to $x_a$, such that $x_a$ belongs to style $s_a$ but has the same semantic content as $x_b$.

\begin{figure*}[tbh]
\includegraphics[width=\textwidth]{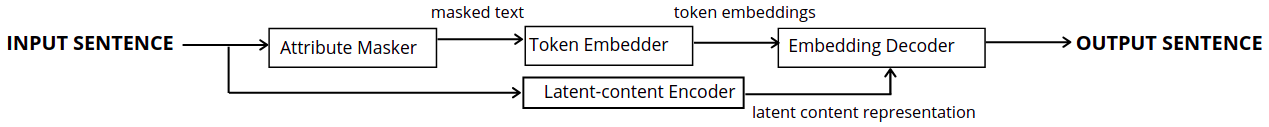}
\caption{The architecture of our proposed model. The model consists of four main components. The arrows show the flow of information within the model, and how the various components interact with each other.} \label{fig:model_architecture}
\end{figure*}

Our model is basically composed of four main components, as illustrated in Fig~\ref{fig:model_architecture}. Before delving into the details, we will summarize the functions of the various components:
\begin{enumerate}
    \item \textbf{Attribute Masker:} The Attribute Masker identifies the attribute words responsible for the bias in a text and masks these words with a special \textit{[MASK]} symbol. The resultant text is fed as input to the Token Embedder.
    \item \textbf{Token Embedder:} The Token Embedder is responsible for generating token embeddings for the masked tokens. These embeddings should contain enough semantic information from the context tokens.
    \item \textbf{Latent-content Encoder:} The Latent-content Encoder takes the original (unmasked) text as input and encodes it into a latent content representation. An important part of this stage is our approach to disentangle the resulting latent content representation from the biased style.
    \item \textbf{Token Decoder:} The Token Decoder computes the average of each token embedding and the latent content representation to generate new token embeddings. The Token Decoder uses these embeddings to predict the correct tokens. 
\end{enumerate}

\begin{figure*}[tbh]
\includegraphics[width=\textwidth]{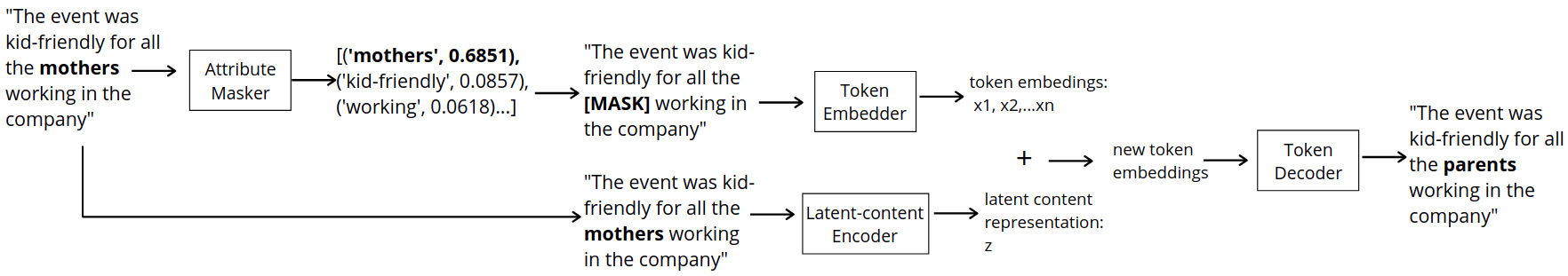}
\caption{An example to illustrate the end-to-end bias mitigation process. This demonstrates the operation of each component of the model. The Latent-content Encoder is debiased to remove traces of bias before being added to the token embeddings.}
\label{fig:example1}
\end{figure*}

We illustrate the process described above with an example in Fig.~\ref{fig:example1}.

\subsection{Attribute Masker}
We use LIME \cite{ribeiro2016should}, a model agnostic explainer that can be used on textual data, to identify attribute tokens. We found LIME to produce good and intuitive explanations, making it an ideal tool for identifying attribute tokens. Some Explicit Style Keyword Replacement models use relatively simple techniques to identify attribute words. \citet{li2018delete}  use the relative frequency of words in the source style. Others like \citet{sudhakar2019transforming} employ more advanced methods like using attention weights.
However, using techniques like attention weights to identify attribute tokens has proven not very effective \cite{jain2019attention}.

To use LIME to detect attribute words, we first need to train a text classifier $f$ that predicts whether a given text is biased. We fine-tune BERT \cite{devlin2018bert}, a pretrained language model, as a text classifier by training it on a labeled corpus containing both biased and neutral texts. 
Lime linearly approximates the local decision boundary of $f$ and assigns weights to tokens based on their influence on the classification outcome.
With these weights (scores), we set a threshold value $\mu$ to select words to be masked. These words are replaced by a special \textit{[MASK]} token. 

\subsection{Token Embedder}
Given the text (with masked attribute tokens) $x_b^\prime$, the Token Embedder learns to generate a vector representation $w_i\in \mathbb{R}^{ d}$ for each token $t^{\prime_i} \in x_b^{\prime}$, such that $w_i$ is able to capture semantic information of $t^\prime_i$. The Token Embedder outputs a set of all token embeddings $W=\{w_1,...,w_n\}\in \mathbb{R}^{n\times d}$. Following the convention used by \citet{devlin2018bert}, we take the size of every embedding to be $d=768$ throughout this paper.

\subsubsection{Training the Token Embedder}
We fine-tune a BERT model to generate token embedding using Masked Language Modeling objective by randomly masking about twenty percent of the words in the training corpus. The goal is to predict the masked words based on the context tokens (unmasked words). A classification layer is appended to the BERT embedder such that the BERT embedder learns to generate embeddings that can be correctly decoded by the classification layer; this same classification layer will later be used as the Token Decoder. This process is similar to the original Masked Language Modeling training objective used to pretrain BERT.

Formally, given the input text $x^\prime_b$, the set of masked tokens in the text $T_{\Pi}$, and the set of context tokens in the text $T_{-\Pi}$, the goal is to predict each masked token based on the unmasked tokens. The MLM objective is to minimize:
\begin{equation}
 \mathcal{L}_{mlm} = -\sum\limits_{i=1; t_{\pi_i} \in T_\Pi }^klogP(t_{\pi_i}|T_{-\Pi})   
\end{equation}

\subsection{Latent-content Encoder}
The Latent-content Encoder is responsible for generating a latent content representation of the input sentence.  The Latent-content Encoder takes as input the original text (unmasked) $x_b$ and generates a target latent representation  $\hat{z}$.

After taking $x_b$ as input, the Latent-content Encoder first generates token embeddings $v_i\in \mathbb{R}^{d}$ for each token $t_i \in x_b$. The set of token embeddings $V=\{v_1,...,v_n\} \in \mathbb{R}^{n\times d}$ is mean-pooled to generate $\hat{z} \in \mathbb{R}^{ d}$. Since we want $\hat{z}$ to have the same content as $x_b$ but not the bias that exists in $x_b$, we use a dual objective training to debias $\hat{z}$.

\begin{figure}[tbh]
\centering
\includegraphics[width=\columnwidth]{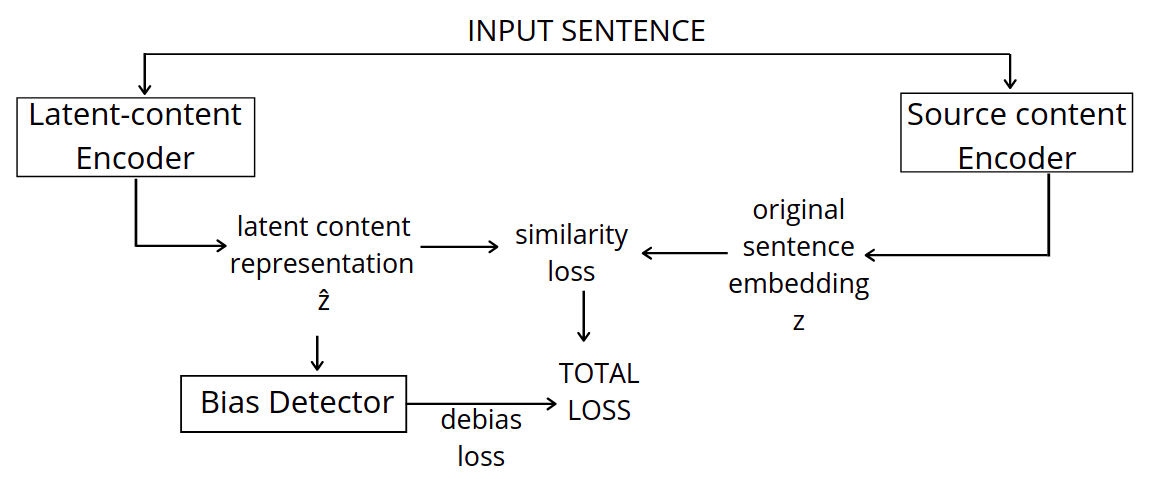}
\caption{Illustration of of the dual objective used to train the Latent-content Encoder. The Source Content Encoder and the bias classifier are both pretrained and frozen during training}
\label{fig:latent_content_encoder}
\end{figure} 

\subsubsection{Training the Latent-content Encoder}
This unique training approach helps to effectively learn a latent content encoding that captures significant content information whilst being disentangled from the biased style.
As shown in Fig~\ref{fig:latent_content_encoder}, a pretrained Source Content Encoder and a pretrained Bias Detector are used to train the Latent-content Encoder.
The Source Content Encoder is pretrained to generate a source encoding $z$ that contains both the content and biased style of $x_b$. The Bias Detector is pretrained to predict the style attribute ($s_b$ or $s_a$). 

Both the Latent-content Encoder and the Source Content Encoder take $x_b$ as input. The Latent-content Encoder generates output $\hat{z}$ whereas the Source Content Encoder generates $z$. Since the goal is to make $\hat{z}$ and $z$ have the same content, we want them to be as similar as possible. We use cosine-similarity as a similarity measure to compute this similarity. The similarity loss is minimized using mean-squared error as defined below:
\begin{equation}
    \mathcal{L}_{sim} = \frac{1}{N} \sum\limits_{j=1}^N(cosine\_similarity(\hat{z}_{j},z_j)-1)^2
\end{equation}

The Bias Detector is a classifier used to detect whether bias exists in $\hat{z}$. The Bias Detector takes $\hat{z}$ as input and returns the class probabilities of $\hat{z}$. Because we want $\hat{z}$ to belong to the neutral class, the Latent-content Encoder has to learn to generate $\hat{z}$ that is always classified as neutral. This is achieved by minimizing the  cross-entropy loss:
\begin{equation}
 \mathcal{L}_{acc_{\hat{z}_j}}=-\sum\limits_{j=1}^NlogP(s_a|\hat{z}_j)
\end{equation}
$P(s_a|\hat{z}_j)$ is the classifier's prediction of the probability of $\hat{z}$ being neutral.

Combining both losses we get the dual objective:
\begin{equation}
LCE\_loss = (1-\lambda)\mathcal{L}_{similarity} +\lambda \mathcal{L}_{acc_{\hat{z}_j}}
\end{equation}
We introduce the balancing parameter $\lambda$ that controls the degree of influence of each objective.
The Latent-content Encoder, the Source Content Encoder, and the Bias Detector are BERT models fine-tuned for these purposes.

\subsection{Token Decoder}

The Token Decoder first adds latent content information to word embeddings. To do this, the Token Decoder takes as inputs both $W$  from the Token Embedder and $\hat{z}$ from the Latent-content Encoder. For each $w_i\in W$, a new token embedding $\hat{w}_i \in \mathbb{R}^{d}$ is generated by computing the weighted average of $w_i \in \mathbb{R}^{d}$ and $\hat{z} \in \mathbb{R}^{d}$. 
After generating $\hat{w}_i$, the Token Decoder uses it to predict the right token by computing the probability distribution over all the tokens in the vocabulary. 

\subsubsection{Training the Token Decoder}
The Token Decoder is a simple one-layer feed-forward neural network. It is only trained on a corpus of neutral texts since we want the decoder to always predict a neutral token. During training, $z$ is used in place of $\hat{z}$ because we want the decoder to predict the exact tokens in the input text. For each masked token's embedding $\hat{w}_{i_{\Pi}}$, the goal is to predict the original token $t_{\pi_i}\in T_\Pi$ where $T_\Pi$ is the set of all masked tokens. We compute the decoding loss as:
\begin{equation}
\mathcal{L}_{dec} = -\sum\limits_{i=1; t_{\pi_i} \in T_\Pi}^n log P(t_{\pi_i}|\hat{w}_{i_{\Pi}})
\end{equation}

To further augment this, we use a pretrained classifier to ensure that the output sentence $x_a$ is always neutral. A dual objective is again used in this process:
\begin{equation}
 \mathcal{L}_{acc_{x_a}}=- logP(s_a|x_a)
\end{equation}
\begin{equation}
TD\_loss = (1-\gamma)\mathcal{L}_{dec} +\gamma \mathcal{L}_{acc_{x_a}}
\end{equation}
Where $\mathcal{L}_{acc_{x_a}}$ is the loss from the classifier, and $\gamma$ is a balancing parameter. Because $x_a$ is made up of discrete tokens (one-hot encodings) which are non-differentiable during back-propagation,  we use a soft sampling approach as was done in \cite{wu2019mask,prabhumoye2018style}:
\[ t_{\pi_i}\sim softmax(\textbf{o}_t/ \tau) \]
where $\textbf{o}_t$ is the logits for $t_{\pi_i}$ from the classifier, and $\tau$ is a temperature value that controls the smoothness of the probabilities.  This continuous representation approximates the one-hot vector that is ordinarily used to represent $t_{\pi_i}$.

\section{Experiments}
For our experiments, we focus on gender bias. Nonetheless, our work is fully adaptable to other forms of biases such as racial bias. The use of gender is motivated by the relative availability of resources such as datasets in this domain from previous works.
To further probe the performance of our model, we experiment on gender obfuscation, where instead of mitigating the bias, we try to convert female-authored texts to "look like" male-authored texts.

We discuss the details of our experiments in the following subsections. All experiments are run on a Tesla V100-SXM3 GPU with 32Gb memory. 

\subsection{Dataset}
We run our experiments on two datasets discussed below. 
Some statistics of the datasets are given in Table~\ref{tab:dataset_statistics}

\subsubsection{Jigsaw dataset:}
The Jigsaw datasets\footnote{https://www.kaggle.com/c/Jigsaw-unintended-bias-in-toxicity-classification/data} consists of comments that are labeled by humans with regard to bias towards or against particular demographics. 

Using the value 0.5 as a threshold, we extract all texts with gender (male or female) label $\geq$ 0.5 as the gender-biased class of texts and extract a complementary set with gender labels $<0.5$ as the neutral class. 
We broke down all texts of multiple sentences into single sentence texts.

\subsubsection{Yelp dataset:}
We extract this dataset from the preprocessed Yelp
dataset used by \cite{prabhumoye2018style,reddy2016obfuscating}. This dataset contains short single sentences which we use for author gender obfuscation.

\begin{table}
\centering
\begin{tabular}{p{13mm} p{11mm} p{11mm} p{7mm} p{7mm} p{7mm}}
 \hline
 Dataset & Attributes & Classifier & Train & Dev & Test \\
 \hline
 \hline
 Jigsaw-g&  Sexist & 24K & 32K & 1K &   1K \\
                &   Neutral &24K  & 92K  &  3K & 3K \\  \hline
Yelp-s & Male  & 100K & 100K  &  1K & 1K  \\ 
            & Female  & 100K & 100K  & 1K  &  1K \\ \hline
\end{tabular}
\caption{Dataset statistics}\label{tab:dataset_statistics}
\end{table}

\subsection{Evaluation models and metrics}
To evaluate the performance of our model, we compare it to six other models; Delete-only, Delete-and-retrieve \cite{li2018delete}, B-GST, G-GST \cite{sudhakar2019transforming}, CAE \cite{shen2017style} and BST \cite{prabhumoye2018style}, all of which are discussed in the section under related works. We chose these models primarily because of their wide use in style-transfer literature and their reproducibility.

The evaluation is based on three automated evaluation metrics for style transfer discussed by \citet{hu2020text}; \textit{style transfer accuracy} (Transfer strength), \textit{content preservation}, and \textit{fluency}.
\begin{itemize}
    \item \textbf{Style transfer accuracy:} This gives the percentage of texts that were successfully flipped from the source style (bias style) to the target style (neutral style) by our model. To predict whether a text was successfully flipped, We use a trained BERT classifier with $85\%$ accuracy and $80\%$ accuracy for the Jigsaw and the Yelp datasets respectively. To ensure fair evaluation, these classifiers are different from the ones used to train the respective models.
    \item \textbf{Content preservation:} We measure content preservation by computing the similarity between the generated text and the original text. Similar to \citet{fu2018style}, we use the cosine similarity between the original text embedding and the transferred text embedding to measure the content preservation. To make this more effective, we generate text embeddings with SBERT \cite{reimers2019sentence}, a modified version of pre-trained BERT that generates semantically meaningful sentence embeddings for sentences so that similar sentences have similar sentence embeddings, that can be compared using cosine-similarity.
    \item \textbf{Fluency:} Similar to \cite{subramanian2018multiple}, we measure the fluency of the generated text using the \textit{perplexity} produced by a Kneser–Ney smoothing 5-gram language model, KenLM \cite{heafield-2011-kenlm} trained on the respective datasets.
\end{itemize}

\subsection{Hyper-parameter settings}

For the gender mitigation task, we set the hyper-parameters as follows: we use a sequence size of 40. We set the balancing parameter for the Latent-content Encoder at $\lambda=0.5$, the balancing parameter for the decoder at $\gamma=0.3$, and the LIME score threshold for attribute token removal at $\mu>0.1$.

For the gender obfuscation task, we set the hyper-parameters as follows: we use a sequence size of 30. We set the balancing parameter for the Latent-content Encoder at $\lambda=0.5$, the balancing parameter for the decoder at $\gamma=0.7$, and the LIME score threshold for attribute token removal at $\mu>0.1$. 

For both tasks, we set the temperature value to $\tau=1$. Other hyper-parameter values are set to the default parameter values used in BERT-base \cite{devlin2018bert}. We also initialize the weight of the Latent-content Encoder with the weights of SBERT before training it. Both models were trained on 4 epochs.
All the hyper-parameter were obtained after experimenting with some hyper-parameter values. We leave a comprehensive discussion of tuning these hyper-parameters to future works.

\subsection{Results and discussion}

In Table~\ref{tab:evaluation_Yelp}, we evaluate our model against the six other models mentioned earlier for gender obfuscation, and in Table~\ref{tab:evaluation}, we do the same for five of the models (excluding BST) for bias mitigation. For style transfer accuracy (A.C.) and content preservation (C.P.), higher values are better. For PPL, lower values are better. Generally, all models produced better style transfer accuracy scores for the bias mitigation task compared to the gender obfuscation task. This is expected because the task of author gender obfuscation appears to be more difficult and indistinct; in the sense that we are assuming that we can predict the gender of a person based on how a person writes. Hence, changing the style is much more delicate. On the other hand, for the bias mitigation task, the distinction between a gender-biased text and a neutral text is relatively obvious and much clearer; attribute words are more distinct and identifiable, as such, more explicit cues can be learned by the model to resolve the text.

\subsubsection{Gender bias mitigation}
\begin{table}[t]
\centering
  \begin{tabular}{ p{18mm} p{12mm}  p{12mm} p{12mm}}
    \hline

    & C.P.\% & PPL & AC\% \\
    \hline \hline
    Original*  & 100.00  & 12.51 & 0.08\\
    Del  &97.47  & 363.64 & \textbf{92.30}\\
    Del\&ret &97.50  & 242.33 & 71.70  \\
    B-GST & 96.73 &1166.4  & 10.10 \\
    G-GST & {99.11}& 621.50 & 38.80\\
    CAE &  95.60 &795.58  & 83.70 \\
    Our model  & \textbf{99.71} & \textbf{76.75} & 88.10\\
  \end{tabular}
  \caption{\textbf{Jigsaw dataset-} Transfer strength and Content preservation scores for the models on all three datasets.  \textbf{C.P.}: Content preservation, \textbf{PPL}: Fluency (Perplexity), \textbf{Accuracy}: Style transfer accuracy, \textbf{Original*:} refers to the original input text. For A.C., C.P.and Agg, higher values are better. For PPL, lower values are better}\label{tab:evaluation}
\end{table}

From Table~\ref{tab:evaluation}, as we expected from the compared models, the models that perform considerably well in one metric suffer significantly in other metrics. For instance, Delete-Only (Del) produces the best transfer accuracy but lags behind other models in content preservation and fluency.
For content preservation and fluency, our model produces improved results over all the other models. This result is consistent with our expectation of improving content preservation with our techniques. Again, the accuracy score (second highest) produced by our model confirms the claim that our model preserves content information without a significant drop in transfer accuracy.

\begin{table}[t]
\centering
  \begin{tabular}{ p{18mm} p{12mm}  p{12mm} p{12mm} }
    \hline
 
     & C.P.\% & PPL & AC\%\\
    \hline \hline
    Original*   & 100.00 & 11.39 & 17.80\\
    Del & 98.70 & \textbf{41.03}  &33.79 \\
    Del\&ret &98.25 &57.73  &30.90 \\
    B-GST  & 95.94 & 141.81 &23.90\\
    G-GST  & 97.28 & 70.24 &21.00 \\
    CAE & 98.48 &43.78 & 32.09 \\
     BST  & 95.49 & 63.33 & \textbf{68.80}\\
    Our model  & \textbf{99.05} & 45.17 & 43.20 \\
  \end{tabular}
  \caption{\textbf{Yelp dataset- }Transfer strength and Content preservation scores for the models for the . \textbf{C.P.}: Content preservation, \textbf{PPL}: Fluency (Perplexity), \textbf{Accuracy}: Style transfer accuracy, \textbf{Original*:} refers to the original input text. . For A.C., C.P.and Agg, higher values are better. For PPL, lower values are better}\label{tab:evaluation_Yelp}
\end{table}

\subsubsection{Gender obfuscation}
From Table~\ref{tab:evaluation_Yelp}, the same observation is made; models that perform very well in one metric fall short in other metrics. BST produces the best style transfer accuracy whilst at the same time producing the worst content preservation score.

\subsubsection{General observation}
From the results from both datasets, one key observation is that models that perform very well in one metric tend to fall short in other metrics. This goes to show the difficulty for style transfer models to preserve content information whilst maintaining a strong transfer accuracy. This observation is confirmed by previous works \cite{li2018delete,wu2019mask,hu2020text} which mention the general trade-off between style transfer accuracy and content preservation. Our model, however, shows good results in maintaining a good balance across all metrics. 
By adjusting the balancing parameters in the hyper-parameter section, priority can be given to which metric is more desired; since it is the case that depending on the application domain, certain metrics might be prioritized. 

Some text samples from the experiments are shown in Table~\ref{tab:sample_texts}.

\begin{table}[t]
\centering
  \begin{tabular}[t]{p{30mm} p{10mm} p{10mm} p{10mm} p{10mm} p{10mm} p{10mm} }
    \hline 
    & C.P.\% & PPL & ACC\% \\
    \hline \hline
    Our model & \textbf{99.05} & 45.17   & 43.20   \\
    Without-LR  & {96.62} & 45.72 & \textbf{84.20}  \\
    Without-LR\&SS  & 96.89 & \textbf{41.84}  & 41.00 \\
    \hline

  \end{tabular}
  \caption{Ablation study of our model on the Yelp dataset. \textbf{Without-LR}: model with soft sampling (class constraint) but no latent content representation, \textbf{Without-LR\&SS}: model with no class constraint and no latent content representation }\label{tab:ablation_study}
\end{table}

\subsubsection{Ablation analysis}

In Table~\ref{tab:ablation_study}, we perform an ablation analysis of our model. Here, we strip off various components from our model and we evaluate the resulting model on the Yelp dataset. We generate two resulting variants; Without-LR and Without-LR\&SS. In Without-LR, we remove latent content information from our model but maintain class constraint ($\mathcal{L}_{acc_{x_a}}$), whereas in Without-LR\&SS both class constraint and latent content information are omitted. 

In Table~\ref{tab:ablation_study}, we can see how each component affects the performance of our model.
We expect Without-LR to have the best style transfer accuracy but the worst content preservation score, for one major reason; class constraint without latent content encoding will ensure that the generated text is mostly in the target style (neutral style) but the loss of content information from the latent encoding reduces content preservation. The exceptionally high transfer accuracy scores produced can be misleading without taking content preservation into account. This is because, in most cases, Without-LR learns to replace female-attribute words with the same word, "pistol". This changes the style in almost all cases but gives the text little meaning. This explains the low content preservation score as well. Since only a few attribute words are changed for a particular text instance, a content preservation score of 96.62\% is significantly low, indicating a significant loss of content information from the replaced word. 
On the other hand, Without-LR\&SS produces the best fluency score primarily because without the class constraint, the model prioritizes predicting a plausible replacement rather than a replacement that changes the style.

\begin{table}[]
\footnotesize
\centering
  \begin{tabular}[]{p{13mm}|p{62mm}}
    \hline
      \multicolumn{2}{c}{\textbf{Gender bias mitigation (biased $\rightarrow$ neutral): Jigsaw dataset}} \\
      \hline
          \hline

      input text
            & i hope the \textit{man} learned his lesson to slow down and buckle up .   \\ \hline
      our model
      & i hope the \textit{driver} learned his lesson to slow down and buckle up .\\
        \hline \hline
        
    input text
    & i married a wonderful mature , loyal and dedicated foreign \textit{women} while working abroad ...  \\ \hline
      our model
      & i married a wonderful mature , loyal and dedicated foreign \textit{person} while working abroad ...\\
    \hline
    \hline
      \multicolumn{2}{c}{\textbf{Gender obfuscation (female $\rightarrow$ male): Yelp dataset}} \\
    \hline \hline
     input text & overall , worth the \textit{extra} money to \textit{stay} here . \\
    \hline
    our model &  overall , worth the \textit{damn} money to \textit{eat} here . \\
    \hline \hline
      input text
      &  i had prosecco and my \textit{boyfriend} ordered a beer . \\ \hline
      our model
      & i had prosecco and my \textit{wife} ordered a beer .\\
  \end{tabular}
  \caption{Sample text outputs from experiments}\label{tab:sample_texts}
\end{table}

\section{Conclusion}
In this work, we introduce a style transfer model that can be used to mitigate bias in textual data. We show that explicit keyword replacement can be effectively combined with latent content representation to improve the content preservation of text style transfer models.
We show the efficacy of our model on bias mitigation as well as other style transfer tasks like gender obfuscation by evaluating the performance on three style transfer metrics; content preservation, fluency, and style transfer accuracy. On both datasets, our approach produces good content preservation scores and fluency scores whilst maintaining a good style transfer accuracy. The key novelties of our approach lie in the combination of explicit keyword replacement and latent content information, our approach of using a text explainer to identify attribute tokens, and our method of generating latent content representation.

As part of our future work, we intend to expand this work to other languages. We plan to explore possible improvements to the model with inclusions such as adversarial learning.
Again, we intend to investigate other forms of attributes beyond tokens, such as sentence length, and how that affects bias in textual data. We also plan apply our model as a preprocessing technique to train fair language models. We believe this could significantly reduce biases found in automated language systems.

\section{Acknowledgement}
This research was supported by the Flemish Government under the “Onderzoeksprogramma Artificiele Intelligentie (AI) ¨
Vlaanderen” programme.

\bibliography{aaai22}

\end{document}